\documentclass[10pt,twocolumn,letterpaper]{article}

\usepackage{iccv}
\usepackage{times}
\usepackage{epsfig}
\usepackage{graphicx}
\usepackage{amsmath}
\usepackage{amssymb}
\usepackage{subcaption}
\usepackage[accsupp]{axessibility}  

\usepackage[pagebackref=true,breaklinks=true,letterpaper=true,colorlinks,bookmarks=false]{hyperref}

\iccvfinalcopy 

\def\iccvPaperID{7646} 

\ificcvfinal\fi

\begin{document}

\title{Keypoint Communities}

\author{Duncan Zauss, Sven Kreiss, Alexandre Alahi \\
EPFL VITA lab \\
CH-1015 Lausanne\\
{\tt\small duncan.zauss@epfl.ch}
}

\maketitle
\ificcvfinal\thispagestyle{empty}\fi

\begin{abstract}

We present a fast bottom-up method that jointly detects over 100 keypoints on humans or objects, also referred to as human/object pose estimation. 
We model all keypoints belonging to a human or an object --the pose-- as a graph and  leverage insights from community detection
to quantify the independence of keypoints.  We use a graph centrality measure to assign
training weights to different parts of a pose.
Our proposed measure quantifies how tightly a keypoint is connected to its neighborhood. 
Our experiments show that our method outperforms all previous methods for
human pose estimation with fine-grained keypoint annotations on the face, the hands
and the feet with a total of 133 keypoints.
We also show that our method generalizes to car poses.
\end{abstract}

\section{Introduction}
Recent large-scale datasets with fine-grained annotations of complex poses
present a new challenge for pose estimation methods. Beyond detecting a coarse
person bounding box and a small set of keypoints for large body joints, we now
have large datasets that include over 100 extra fine-grained keypoints in the
face, the hands and the feet. Resolving
these fine details will allow us to build robust representations of humans
for downstream tasks like action recognition \cite{Mokhtarzadeh2019cvpr,Bertoni2020sociald}, and intent prediction \cite{Mordan20its, haziq2020}.

\begin{figure}[t]
  \centering
  \includegraphics[width=1.0\linewidth,trim=0cm 0cm 1.29cm 0cm,clip]{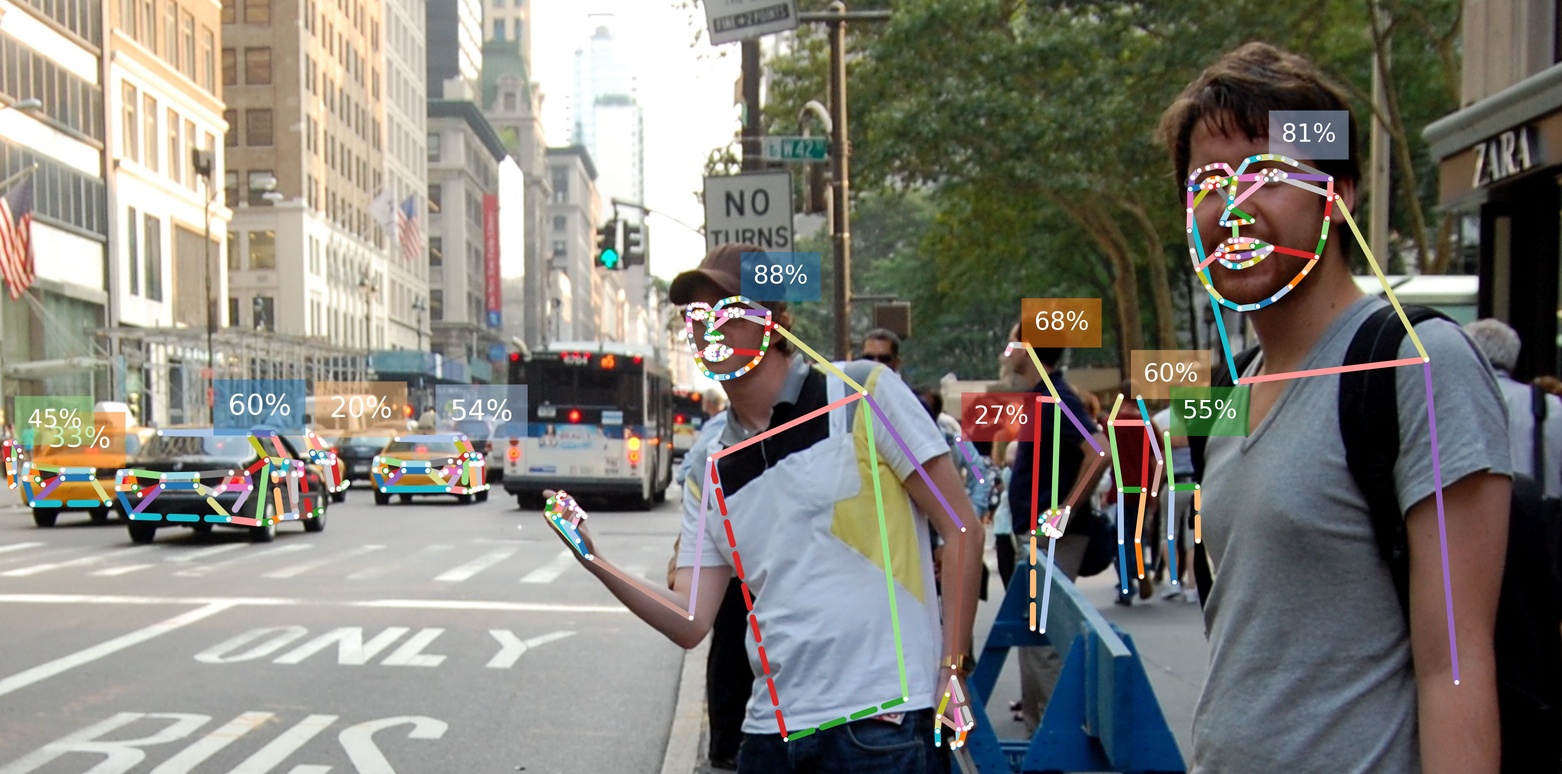}
  \caption{
  Our proposed keypoint weighting method for complex poses yields state-of-the-art results for whole body human pose estimation and for complex car poses, whilst running at high frame rates.
  }
  \label{fig:pull}
\end{figure}

Training our current pose estimation algorithms on poses with mixed coarse
and fine keypoints presents a challenge as they assume a uniform importance
of all keypoints in a pose.
We introduce a principled keypoint weighting method to take into account
the difference of the importance of coarse and fine-grained keypoints.
Figure~\ref{fig:pull} shows a complex person and car poses. 
For instance, the person pose contains coarse keypoints like the hips and
shoulders and fine-grained keypoints like the ones along the eyebrows.


In~\cite{jin2020whole}, Jin \etal share a large scale annotation for complex
human body poses and propose their method, ZoomNet, that set the state-of-the-art
for this type of complex human body pose. Their method first localizes a person
with their major keypoints and then estimates the areas of the hands and face.
On the estimated areas, it runs a separate head that is zoomed-in on that area
to determine fine-grained keypoint locations. In contrast, we propose a fast,
bottom-up
method that directly estimates all keypoints in parallel. Our method does
not need predefined areas for fine-grained estimation and, therefore, generalizes
to any pose, like a fine-grained car pose. Such a car pose is proposed in the
ApolloCar3D dataset~\cite{song2019apollocar3d} and we show that our method
generalizes to this pose as well.

Complex, coarse and fine-grained poses present a challenge for current pose
estimation methods that assume a uniform distribution of keypoints across
a person or object. A cluster of fine-grained keypoints overly emphasizes that
region and focuses the neural network optimization on that area reducing the
importance of other regions that only have a single keypoint. We propose a
method that quantifies how tightly connected these keypoints are and that rebalances
the training weights such that all areas of a pose are equally well connected
to the rest of the pose. We introduce the details in Section~\ref{sec:method}.

Our contributions are
(i)~a method to weigh the importance of keypoints and their connections in
complex poses based on graph-based methods for community detection,
(ii)~an efficient implementation for fine-grained human poses and
(iii)~generalization from human poses to fine-grained car poses.
We show the impact of our contribution with state-of-the-art
results on the challenging COCO WholeBody dataset~\cite{jin2020whole}
and the ApolloCar3D dataset~\cite{song2019apollocar3d}.
The software is open source and publicly available \footnote{\url{https://github.com/DuncanZauss/Keypoint_Communities}}.

\section{Related Work}

There is an extensive literature on pose estimation. While many works have
focused on human pose estimation, there are recent works
that extend the method to animal pose estimation~\cite{mathis2018deeplabcut}
and car pose estimation~\cite{song2019apollocar3d}. Recent datasets include more keypoints that represent
finer details on human and car poses and are reviewed below.

\paragraph{Human Pose Estimation.}
The recent release of the COCO WholeBody dataset~\cite{jin2020whole}
with 133~keypoints for a single human pose presents new challenges
for existing methods. The authors, Jin \etal, established baseline numbers
of existing methods on their dataset and proposed ZoomNet, a new neural
network architecture that refines regions with fine-grained annotations with
dedicated networks.

In general, state-of-the-art methods for human pose estimation are based on
Convolutional Neural Networks~\cite{toshev2014deeppose,he2017mask,cao2017realtime,newell2017associative,papandreou2018personlab,xiao2018simple,sun2019deep,wei2016convolutional,newell2016stacked,kocabas2018multiposenet,cheng2020higherhrnet}.
There are two major approaches for pose estimation. Bottom-up methods
estimate each body joint first and then group them into poses. Top-down methods
first run a person detector to estimate person bounding boxes before estimating
body joint locations within each bounding box.

The first bottom-up methods were introduced, \textit{e.g.}, by Pishchulin \emph{et al.} with
DeepCut~\cite{pishchulin2016deepcut}.
They solve the keypoint association problem with an integer linear program.
In these early methods, the processing time for a single image was of the order of hours.
Newer methods introduced additional concepts to reduce prediction time, \emph{e.g.},
in Part Affinity Fields~\cite{cao2017realtime},
Associative Embedding~\cite{newell2017associative},
PersonLab~\cite{papandreou2018personlab} and
multi-resolution networks with associate embedding~\cite{cheng2020higherhrnet}.
Composite Fields as introduced in PifPaf~\cite{kreiss2019pifpaf} predict
more precise associations than
OpenPose's Part Affinity Fields~\cite{cao2017realtime} and
PersonLab's mid-range fields~\cite{papandreou2018personlab}
which allows for particularly fast and greedy decoding with high precision.

\begin{figure*}[t]
  \centering
  \includegraphics[width=0.9\linewidth]{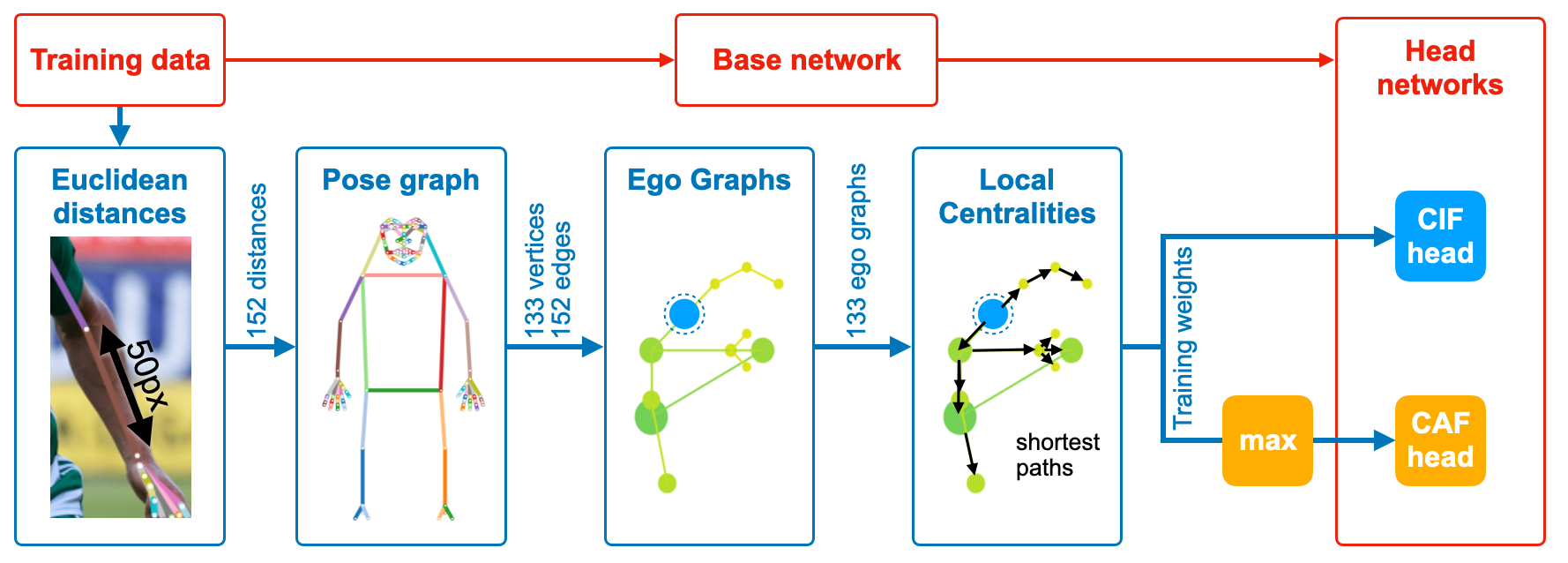}
  \caption{
    Overview of our method. We obtain the average euclidean distance for
    every connection in the pose graph from the training dataset.
    We then create ego graphs of radius three for every vertex and compute the
    local centrality for the ego vertex. The centrality is directly related
    to the training weight of a joint. The training weight of a connection
    is obtained by taking the max of the weights from the vertices of this
    connection.
  }
  \label{fig:method}
\end{figure*}

\paragraph{Car Pose Estimation.}
The new ApolloCar3D dataset~\cite{song2019apollocar3d} with its 66~keypoints
 for car pose estimators presents similar challenges than the WholeBody dataset~\cite{jin2020whole}. The authors~\cite{song2019apollocar3d} presented
baseline performance numbers using Convolutional Pose Machines (CPM)~\cite{wei2016convolutional}
and also quantified the performance of human labelers on their dataset.

It is only recently that methods that were developed for human pose estimation
have been applied to other classes.
Car poses provide finer detail for a car than a 2D or 3D
detection bounding box. While human pose estimation focuses on the location
of body joints within the human body, car poses annotate points on the surface
of the car.

One of the earlier works by Reddy~\etal proposes
Occlusion-Net~\cite{reddy2019occlusion} that highlights the issue of self-occlusion
for these keypoints on the surface of an object. As the car is viewed from
different sides, the set of visible keypoints changes drastically due to
self-occlusion. Their work includes extensive modeling with a 3D graph network
and self-supervised training with the CarFusion dataset~\cite{dinesh2018carfusion}
to predict 2D and 3D keypoints.
In OpenPose~\cite{cao2019openpose}, Cao~\etal show qualitative results for
car pose estimation.
Simple Baseline~\cite{sanchez2020simple} trains a top-down pose estimator on
car annotations of the Pascal3D+ dataset~\cite{xiang2014beyond}.
Other works choose different representations for finer details beyond bounding
boxes. In GSNet~\cite{ke2020gsnet}, a car orientation in 3D space is predicted along
with a 3D shape estimate.

\paragraph{Keypoint weighting}  
In general, previous methods used uniform distributions to weigh the keypoints in training for single networks or used separate networks for the different fine-grained regions. We show that with our keypoint weighting method the performance of single neural networks for predicting poses that contain fine-grained and coarse features can be improved significantly.

\section{Method}
\label{sec:method}

We need to devise a training procedure for poses that combine coarse keypoints
that localize large body parts (hips, shoulders, \etc) and fine-grained keypoints
like the outline of a hand. Individual fine-grained keypoints are highly
predictable from neighboring keypoints whereas coarse keypoints are more
independent. However, the importance of a keypoint is not only based on its
individual predictability, but also on how it contributes to a local group of
keypoints. In other words, every individual keypoint in a group of five keypoints
might have negligible importance, the group of five keypoints together is
still important.

\begin{figure}
  \centering
  \begin{subfigure}[]{0.45\linewidth}
    \centering
    \includegraphics[height=4cm]{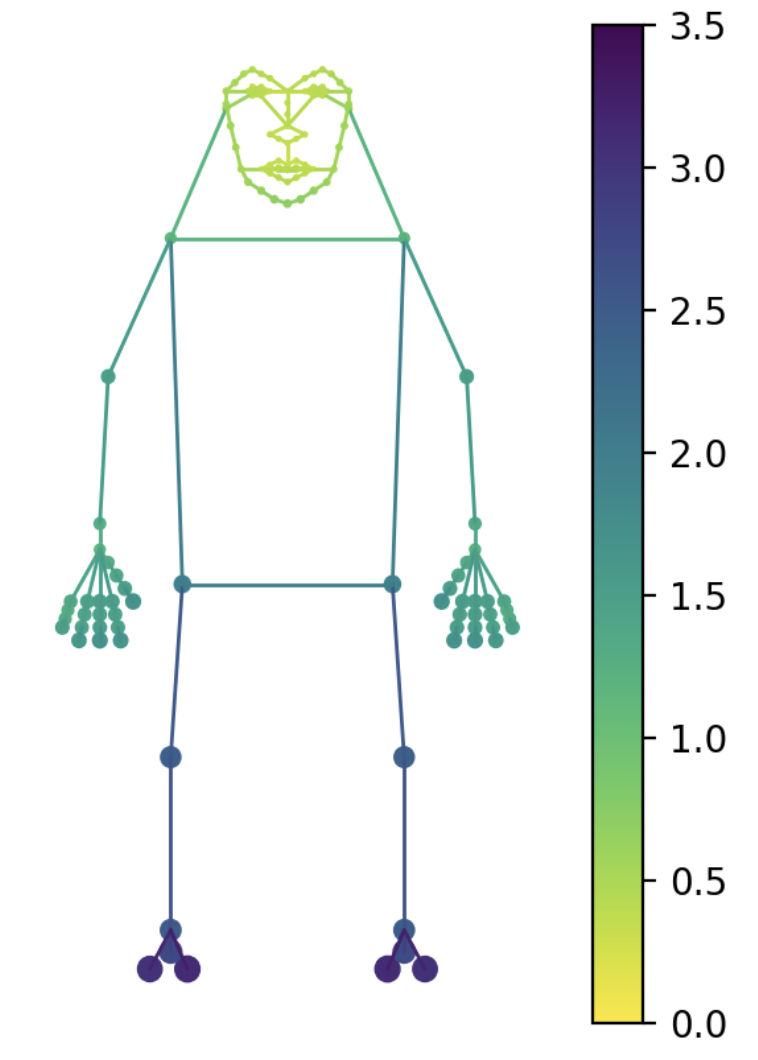}
    \caption{}
  \end{subfigure}
  \begin{subfigure}[]{0.45\linewidth}
    \centering
    \includegraphics[height=4cm,trim=0 1.2cm 0 1.2cm,clip]{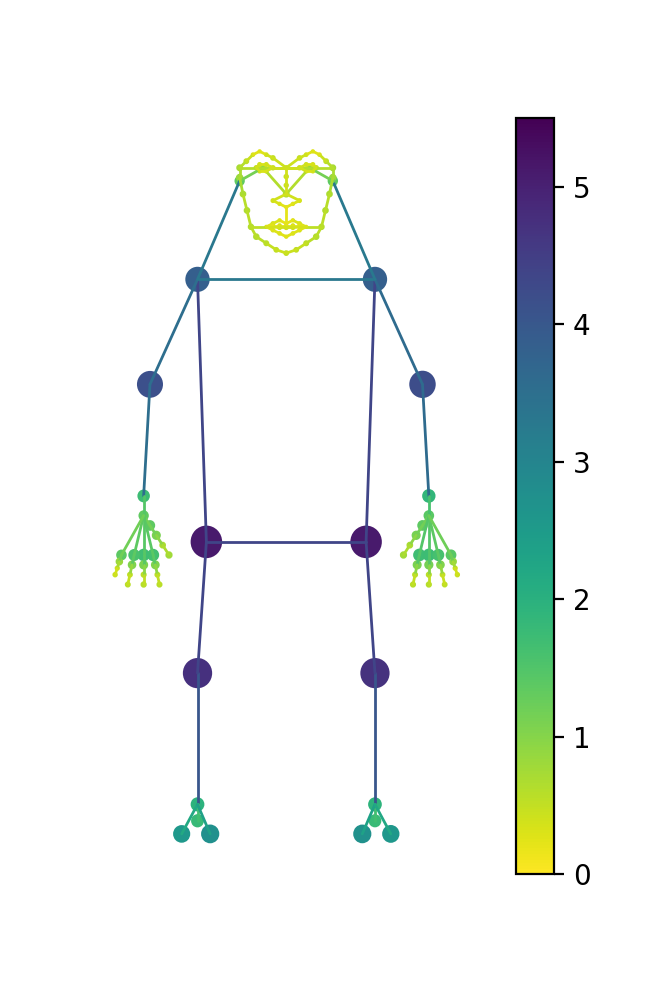}
    \caption{}
  \end{subfigure}
  \caption{
    Visualization of proposed weighting for joints and connections.
    The colors indicate the training weights. For the joints, also the
    circle radius is proportional to the joint weight.
    In~(a), all shortest paths are taken into account. In~(b), only shortest
    paths within a radius of three are taken into account.
  }
  \label{fig:weights-wb}
\end{figure}

\begin{figure*}
  \centering
  \begin{subfigure}[]{0.45\linewidth}
    \centering
    \includegraphics[width=0.85\linewidth]{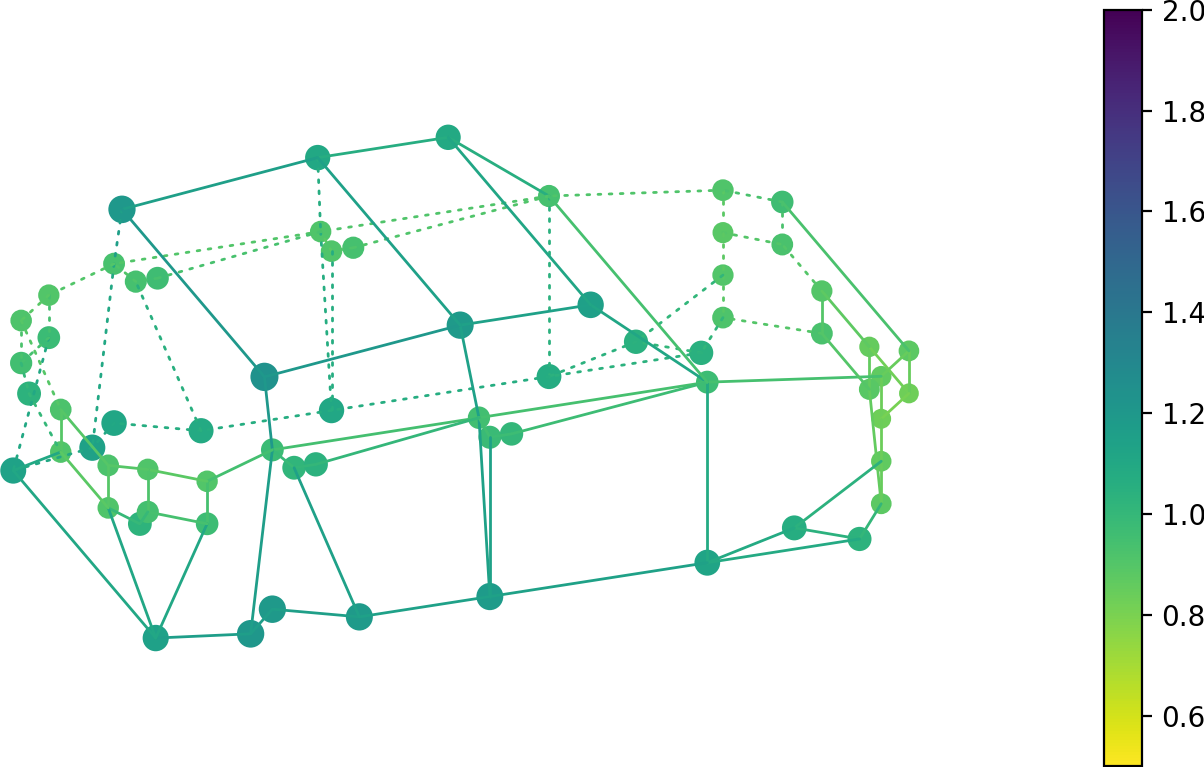}
    \caption{}
  \end{subfigure}
  \begin{subfigure}[]{0.45\linewidth}
    \centering
    \includegraphics[width=0.85\linewidth]{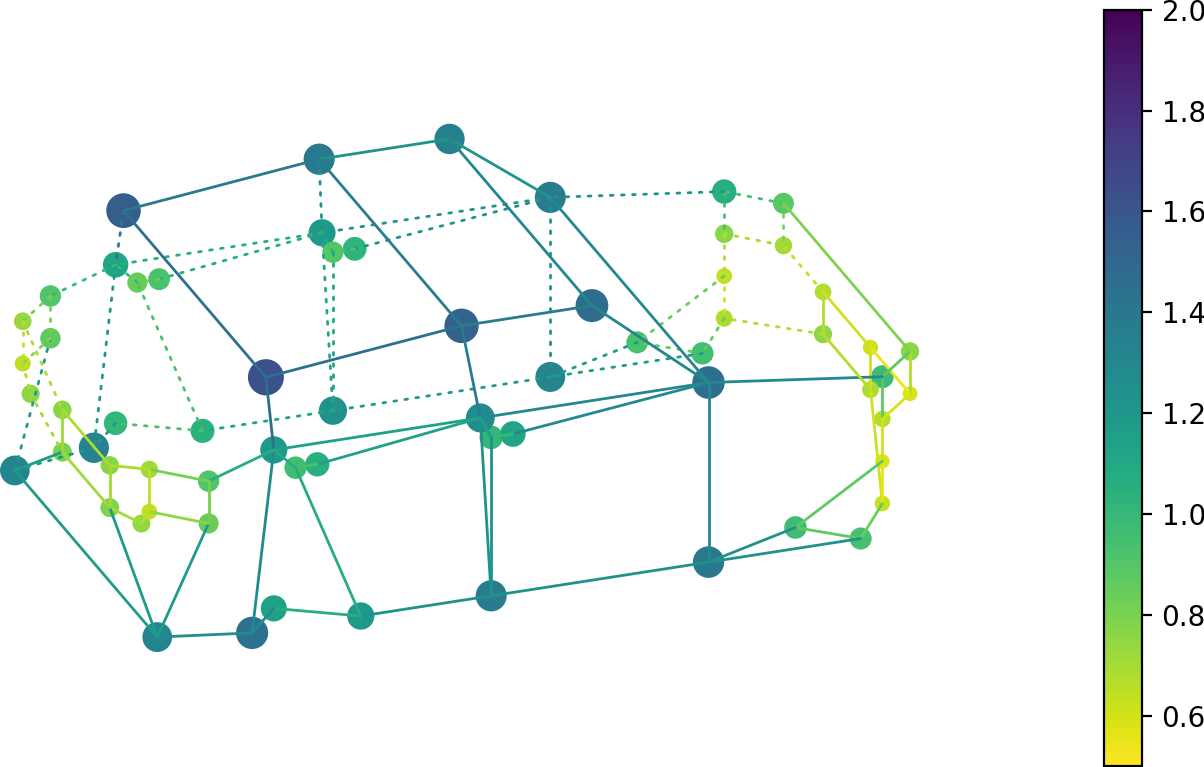}
    \caption{}
  \end{subfigure}
  \caption{
    Visualization of proposed weighting for the skeleton of an instance from
    the ApolloCar3D dataset~\cite{song2019apollocar3d}.
    In~(a), all shortest paths are taken into account. In~(b), only shortest
    paths within a radius of three are taken into account.
    The dotted lines indicate connections on the left side of the car for a
    clearer visualization.
  }
  \label{fig:weights-apollo}
\end{figure*}

\subsection{Pose Estimation Architecture}

Figure~\ref{fig:method} provides an overview of our architecture. Our method is
independent of the particular choice of pose estimator, and could be used for any top-down or bottom up pose estimation algorithm. We will use
OpenPifPaf~\cite{kreiss2019pifpaf}, which is a bottom-up pose detector based on
Composite Fields. A backbone in form of a ResNet~\cite{he2016deep} or
ShuffleNetV2~\cite{ma2018shufflenet} processes single images to create a
common representation for the head networks. The head networks are the
Composite Intensity Fields (CIFs) and Composite Association Fields (CAFs)
that are $1 \times 1$ convolutions followed
by subpixel convolutions~\cite{shi2016real}.
The heads are trained to detect keypoints and
to associate keypoints respectively.
Per keypoint type $k$ and for every location $i,j$ in the output field, the CIF head predicts an intensity component $c^{i,j}_k$ to indicate a
keypoint is
nearby, a two-dimensional vector component $v^{i,j}_k$ to precisely regress to the keypoint location, the uncertainty of the location $b^{i,j}_k$ and
a scale component $s^{i,j}_k$ to estimate the size of a keypoint. The learnt scale of the keypoint  $s^{i,j}_k$
depends on the size of that specific joint in the image and is used in the decoding step as the width of an
unnormalized Gaussian convolution to create high-resolution confidence fields.
Similarly, the CAF head
also has an intensity component to indicate the vicinity of an association
between two keypoints, two vector components that regress to the two keypoints instances
to associate, and two scale components to estimate the two keypoint sizes.
The CIF loss with an extension to weigh all loss components by keypoint
type $k$ with $w_k$ is:
\begin{eqnarray}
  \mathcal{L}_{\textrm{CIF}} &= \sum\limits_k w_k \Bigg[
  & \sum_{m_{k,c}}
      \textrm{BCE}(c, \hat{c})
  \label{eq:loss-confidence} \\
  &&+ \; \sum_{m_{k,v}} \textrm{Laplace}(v, \hat{v}, \hat{b}) \;\;\;\;\;
  \label{eq:loss-localization} \\
  &&+ \; \sum_{m_{k,s}}
    \textrm{Laplace}\bigg(1, \frac{\hat{s}}{s}, b_s \bigg) \;\;\; \Bigg]
  \label{eq:loss-scale}
\end{eqnarray}
where $c$, $v$, $b$ and $s$ are components of the composite field
with suppressed indices ($k, i, j$) for the keypoint type and feature map location and
where symbols with a hat indicate predicted quantities.
With $m_{k,c}$, $m_{k,v}$ and $m_{k,s}$, we indicate keypoint specific masks over the
feature map. BCE is a binary cross entropy loss with Focal loss
extension~\cite{lin2017focal} and Laplace is a linear regression loss for
vector components that is attenuated by a predicted $\hat{b}$ or a fixed
$b_s$~\cite{kendall2017uncertainties,kreiss2021openpifpaf}.
The probabilistic interpretation of the loss function as the negative logarithm of a joint likelihood function requires that the three components of the loss are equally weighted with respect
to each other. The CAF head is trained with an equivalent loss with
two vector components~(\ref{eq:loss-localization}) and two scale components~(\ref{eq:loss-scale}).
We now focus our attention on the blue branch of Figure~\ref{fig:method}
that determines the training weights $w_k$ for the CIF and CAF heads.

\subsection{Graph Centrality}

We represent a pose as a graph $\{V, E\}$ with vertices $V$ representing each keypoint, and edges $E$
representing the Euclidean distance in the image plane between the keypoints.
That Euclidean distance is estimated with an average over all training
annotations. The importance of a keypoint does not only depend on its direct
neighbors, but on its connectedness in the neighborhood. Therefore, we
consider graph centrality measures.

There exists an enormous number of graph centrality measures, for example 
closeness centrality \cite{bavelas1950communication}, eigenvector centrality 
\cite{bonacich1987power}, Katz centrality \cite{katz1953new}, betweenness centrality 
\cite{freeman1977set} and harmonic centrality \cite{marchiori2000harmony} among others.

Traditionally, centrality measures are used to determine the centrality or importance of 
persons in social networks or more generally the importance of nodes in complex graphs. 
Highly central nodes get assigned a high centrality value. For our application highly 
central nodes are part of a community and thus we use the inverse of the centrality.

We want to use
our understanding of the particular problem do derive the best metric for
our use case. To train complex poses, we aim to make every keypoint equally
well connected to the rest of the pose. Most centrality measures are based on
shortest paths from a vertex to all other vertices. The ``connectedness''
of a vertex is represented in the average length of the shortest paths that
originate at the vertex. For example, the ankle keypoint is not very well
connected and the average length of all shortest paths that originate at
the ankle is high as all the paths to the face and hand keypoints are long.
To rebalance our training such that all vertices are connected equally well,
we want to assign an equal weight to an average unit of distance in the
shortest paths.

In practice, we restrict the centrality
computation to a neighborhood by extracting an ego graph of radius three
(the subgraph with all the vertices around a particular vertex that can be
reached in three steps)
for every vertex and computing the centrality for that vertex only within that
subgraph.

The weighted length (weighted by Euclidean distance) of the shortest path
between two vertices $v_1$ and $v_2$
is $d(v_1, v_2)$. As we are interested in equally weighting a unit length,
the harmonic average is appropriate.
For every vertex $v_i$, we compute the harmonic average $h$ of all the shortest
paths originating at $v_i$:
\begin{equation}
  h(v_i) = \left[\sum_{v_j \in V \backslash \{v_i\}} \frac{1}{d(v_i, v_j)} \right]^{-1}
\end{equation}
where we can identify the
harmonic centrality $H$~\cite{marchiori2000harmony} in the square brackets,
leading to $h = H^{-1}$.

Numerically, the closeness centrality~\cite{bavelas1950communication} and
harmonic centrality are similar
and it might be helpful to interpret this weighting in terms of assigning
a high weight to keypoints with low ``closeness''.

\subsection{Training Weights}

We use the graph centrality measure to derive training weights for keypoints
and their connections. The keypoint weights are obtained directly from the
centrality by normalizing $\sum_i h(v_i)$ to the number of keypoints. This
normalization facilitates easier comparisons between the different
weighting methods.

We also need to obtain training weights for keypoint connections.
Again, the weight of the
connection should not depend just on its own length, but also take into account
the structure of the local cluster of keypoints this edge is a part of. Given
we already have a principled method for the vertices, we derive the weight
$w_{ij}$ for the edge that connects vertices $v_i$ and $v_j$ from $h$:
\begin{equation}
  w_{ij} \propto \max \big( h(v_i), h(v_j) \big) \;\;\; .
\end{equation}
We normalize the sum of edge weights to the total number of edges.

The resulting weights for the human WholeBody pose~\cite{jin2020whole} and the car pose~\cite{song2019apollocar3d} are
shown in Figures~\ref{fig:weights-wb} and~\ref{fig:weights-apollo}.
We show two configurations of our method. One where we use the entire
pose to compute our graph centrality measure and one where we use an
ego graph of radius three.
The WholeBody skeleton has clear hierarchical clusters of keypoints in the
hands and face and one level down in the eyes and fingers.
The ApolloCar3D skeleton is more uniformly distributed.
There are keypoint agglomerations in the area
of the lights and number plates both in the front and in the rear of the car. The
keypoints in the roof are the most separate from the other keypoints. In contrast to
the COCO WholeBody skeleton the communities are not as strongly separated.
For the WholeBody pose, the computed weights range from 0.21 to 5.15 and
for ApolloCar3D from 0.57 to 1.63.
Our method automatically determines the keypoint communities for the WholeBody
pose and produces highly varied training weights. Groups of keypoints that
are highly predictable from each other receive a lower weight.
Our method also successfully
determines the more uniform distribution of keypoints in the car pose and
produces less varied training weights.

For any generic pose and training dataset, this method automatically creates
training weights for keypoints and associations in a principled way and
we show its effectiveness on challenging experiments in the next section.






\section{Experiments}

We conduct extensive experiments on complex poses to demonstrate the
effectiveness and efficiency of our method. We investigate a human pose with
17 COCO keypoints as the main skeleton that was then extended with an
additional 116 keypoints for fine-grained details in the face, the hands and
the feet. We also demonstrate that our method generalizes to a fine-grained
car pose with 66 keypoints.

\paragraph{Datasets.}
For human pose estimation, we conduct experiments on the
COCO WholeBody~\cite{jin2020whole} dataset. This dataset contains extra
annotations on the 64,000 training and 5,000 validation images of
COCO~\cite{lin2014microsoft}
for face, hands and feet. The full pose contains 133 keypoints with
152 connections.
There are about 130,000 instances with annotations for the left hand, the right hand and the face.
The body annotations are taken from COCO~\cite{lin2014microsoft} that contains about 250,000 instances.

For car pose estimation, we use the ApolloCar3D
dataset~\cite{song2019apollocar3d}. It provides car annotations with
66~keypoints and we assigned them 108~connections.
The dataset consists of 4283~training and 200~validation images with 52942
and 2674 annotated instances respectively. As cars are only visible from
one side and often partially occlude each other, only an average of
16.2 keypoints are annotated per instance.

\paragraph{Evaluation.}
We follow the evaluation method proposed in the COCO WholeBody~\cite{jin2020whole}
dataset paper. It is based on keypoint-based average precision (AP) that
was popularized with the COCO keypoint task~\cite{lin2014microsoft}.
The evaluation weighs every one of the 133 keypoints equally.

For the ApolloCar3D dataset, we report the detection rate which was proposed by
Song \etal~\cite{song2019apollocar3d}. A keypoint is counted as detected if
the distance from the prediction to the ground truth is less than 10 pixels.
Additionally, we also report the keypoint-based AP as it is
already common for human pose detection~\cite{lin2014microsoft}. We compute
AP based on object keypoint similarity with sigmas of 0.05 for all car keypoints.

\paragraph{Implementation Details.}
We extend OpenPifPaf~\cite{kreiss2021openpifpaf} with an option to weigh
the training of keypoint and connection types. We populate the weights for the
given pose in the generic fashion described in Section~\ref{sec:method}.

We train models with ShuffleNetV2~\cite{ma2018shufflenet} backbones that were
pretrained without weighting the MS COCO keypoint task.
The head networks CIF (Composite Intensity Field) and
CAF (Composite Association Field) are single $1\times1$ convolutions followed
by a subpixel convolution~\cite{shi2016real}. The total stride after the backbone
is 16 and decreased to 8 in the head networks.
We train for
100~epochs with a learning rate of 0.0001 with an SGD~\cite{bottou2010large}
optimizer with Nesterov momentum~\cite{nesterov27method} of 0.95 and a
batch size of~16.

\begin{figure*}
  \centering
  \includegraphics[height=4.6cm,trim=1cm 1cm 1cm 2cm,clip]{{images/coco/000000013291.jpg.predictions}.png}
  \includegraphics[height=4.6cm,trim=0cm 0cm 1cm 2cm,clip]{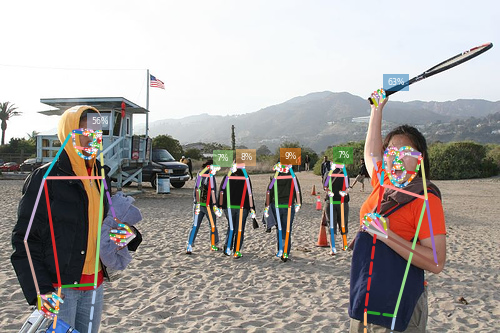}
  \includegraphics[height=5cm,trim=0cm 1cm 1cm 1cm,clip]{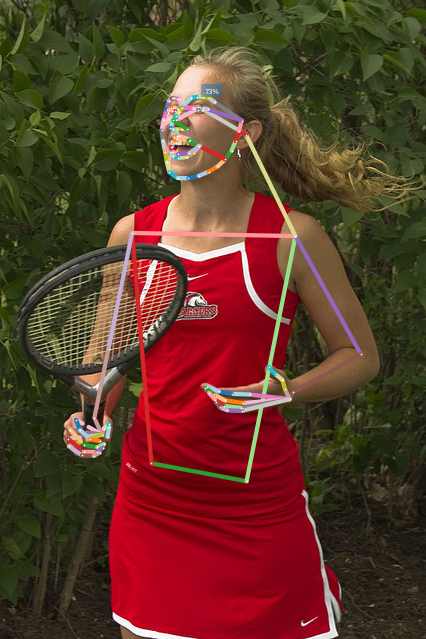}
  \includegraphics[height=5cm,trim=4cm 3cm 4cm 0,clip]{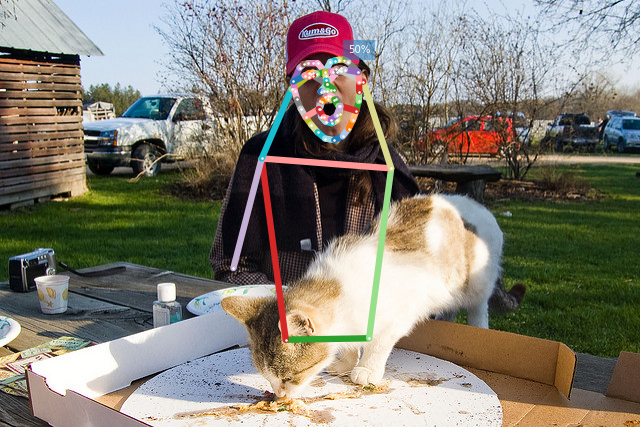}
  \includegraphics[height=5cm,trim=2cm 2cm 2.1cm 0,clip]{{images/coco_emotions/000000334309.jpg.predictions}.png}

\caption{
    Qualitative results from the COCO WholeBody validation set~\cite{jin2020whole}.
    Our method resolves multiple persons per image and captures their
    facial expressions and gestures like hailing a cab.
    The bottom-left image is processed with human and car pose estimators.
  }
  \label{fig:qualitative}
\end{figure*}

\begin{figure*}
  \centering
  \includegraphics[height=4.5cm,trim=3cm 0cm 0cm 2cm,clip]{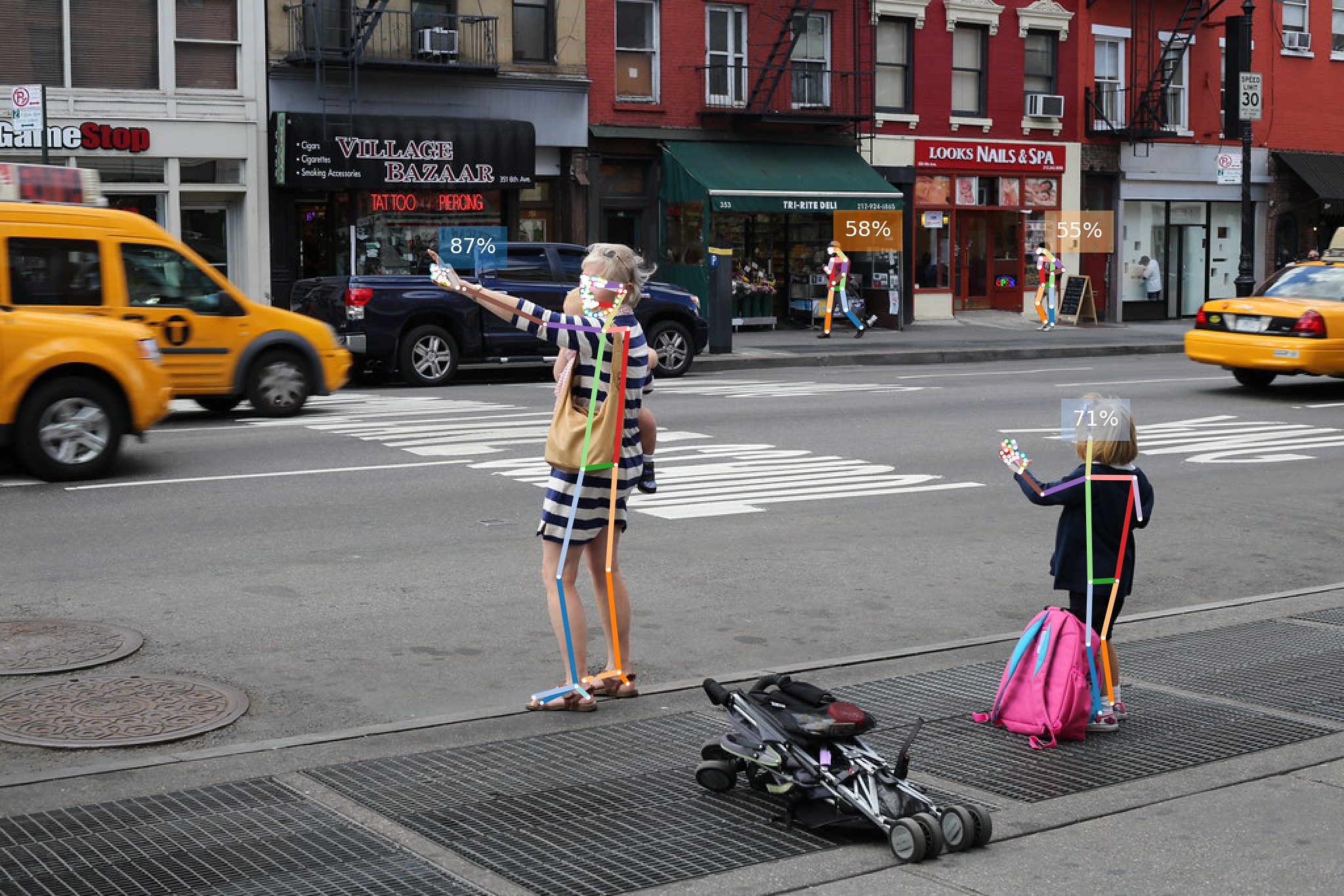}
  \includegraphics[height=4.5cm,trim=0cm 0cm 2cm 2cm,clip]{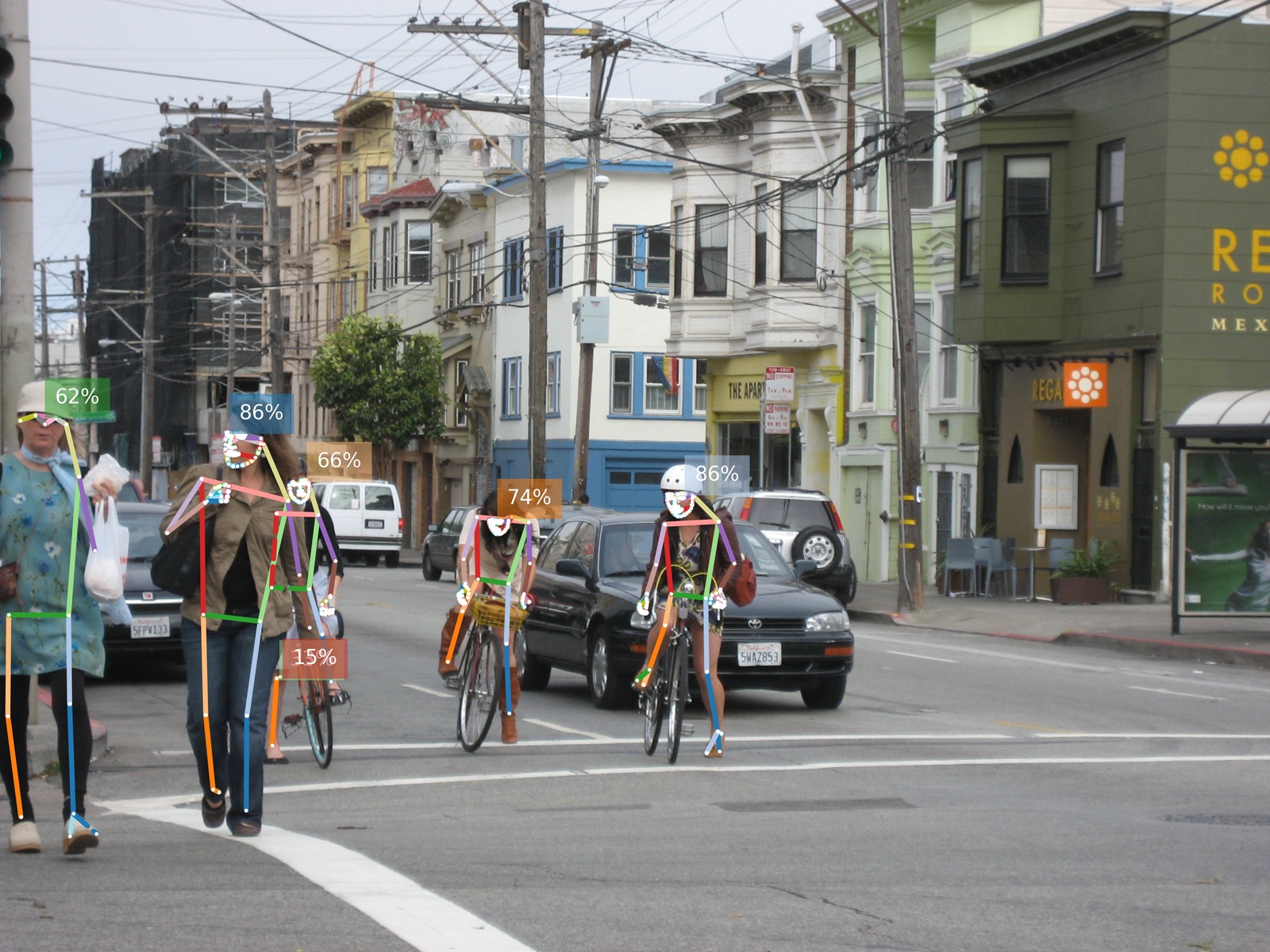}
   \includegraphics[height=4.5cm,trim=0.75cm 1cm 0cm 10cm, clip]{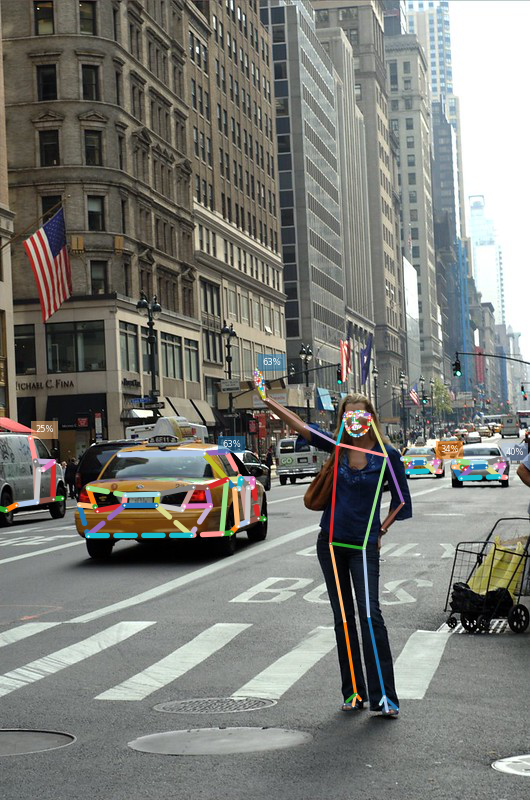}
   \includegraphics[height=4.5cm,trim=0cm 0cm 0cm 0.8cm,clip]{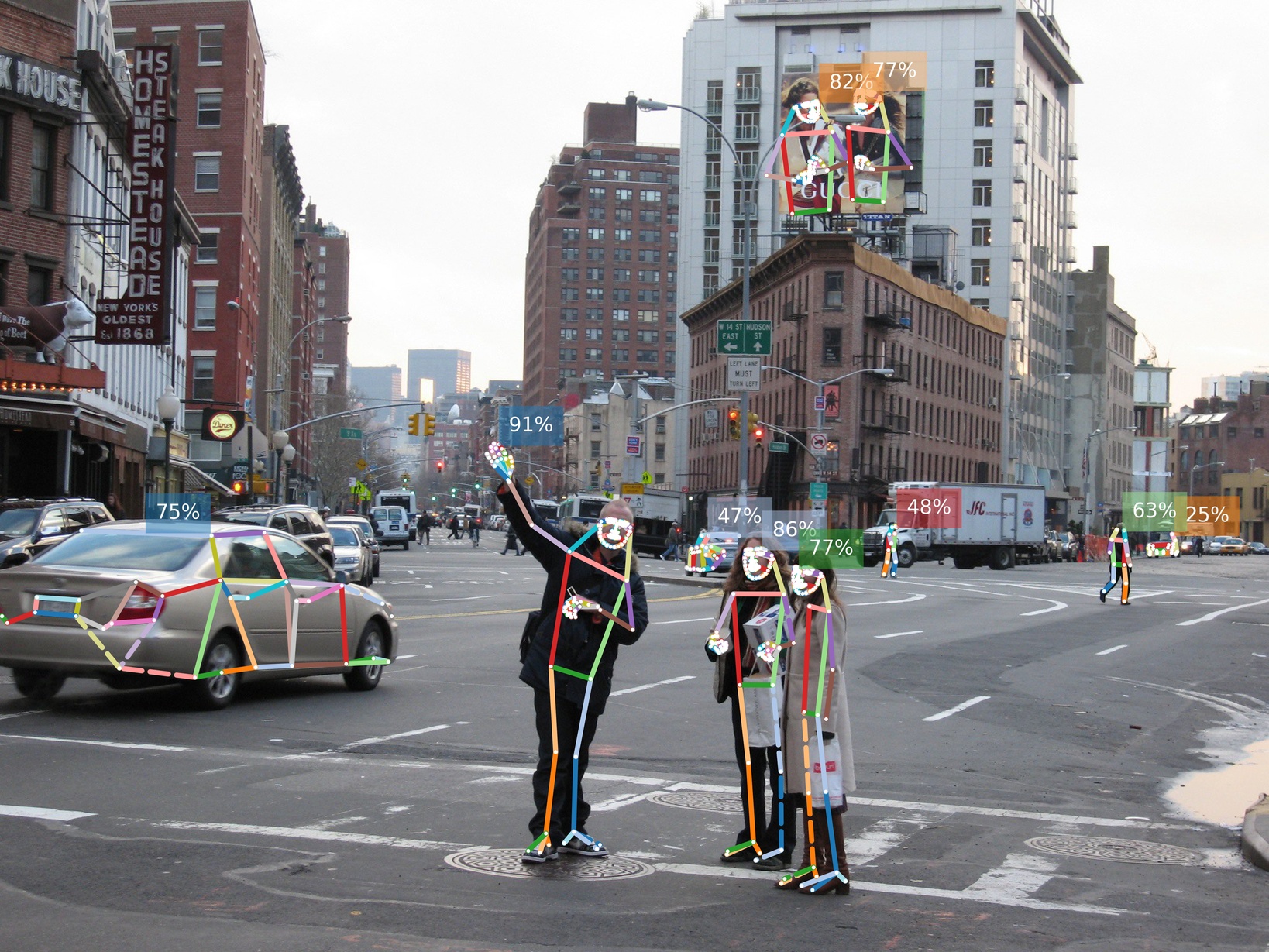}
\caption{
    Qualitative results with images from the transport domain that we obtained from flickr. The two images in the bottom row were processed with our human and car pose estimation network.}
  \label{fig:flickr_qualitative}
\end{figure*}

\begin{table}
  \centering
  \begin{tabular}{|l|c c c c c|}
    \hline
    Method & \textbf{WB} & body & foot & face & hand \\
    \hline\hline
    HRNet~\cite{sun2019deep}        & 43.2 & 65.9 & 31.4 & 52.3 & 30.0 \\
    ZoomNet~\cite{jin2020whole}     & 54.1 & \textbf{74.3} & \textbf{79.8} & 62.3 & 40.1 \\
    \hline
    AE~\cite{newell2017associative} & 27.4 & 40.5 & 7.7 & 47.7 & 34.1 \\
    OpenPose~\cite{cao2017realtime} & 33.8 & 56.3 & 53.2 & 48.2 & 19.8 \\
    \textbf{Ours}    & \textbf{60.4} & 69.6 & 63.4 & \textbf{85.0} & \textbf{52.9} \\
    \hline
  \end{tabular}
  \caption{
    Average precision (AP) results in percent on the COCO WholeBody
    dataset~\cite{jin2020whole}. WB indicates evaluation on all 133 keypoints. The first two methods are top-down methods and the lower three are bottom-up methods.
    Reference numbers from~\cite{jin2020whole}.
  }
  \label{tab:results}
\end{table}

\paragraph{Results on the COCO WholeBody dataset.}
Quantitative results on the COCO WholeBody dataset~\cite{jin2020whole} are
shown in Table~\ref{tab:results}.
Our result is based on a single model that is evaluated for all (WB) or
a subset of the predicted keypoints.
Our method outperforms previous methods and achieves
especially high precision on fine-grained regions such as the face or hands.
The ``body'' task is equivalent to the COCO keypoint task on the val
set~\cite{lin2014microsoft}.
Our method is based on OpenPifPaf~\cite{kreiss2021openpifpaf} which only achieves
71.6\% on the COCO val set with a model trained on that specific task and
we therefore did not expect it to outperform ZoomNet~\cite{jin2020whole}.
Our method makes up for its lower body AP with excellent results for face
and hand AP and is nearly twice as precise as any other bottom-up method.

Qualitative results are shown in Figure~\ref{fig:flickr_qualitative} and Figure~\ref{fig:qualitative}. The additional keypoints in
the face can serve as a powerful representation from which human emotions
such as happiness and surprise but also attention and intent can be derived.
The fine and coarse-grained WholeBody pose can be used to
predict actions from images. For the example image on the bottom right of Figure~\ref{fig:qualitative}, it can be predicted
that the person is eating while holding a cell phone and for the person on the middle picture
in the bottom row it can be predicted that the person is surprised
that a cat jumped on the table.
Through adding fine-grained hand keypoints the interaction between humans and objects can be detected.
In the transportation domain, the fine grained keypoints on the hand region help to understand
the intentions of pedestrians. For example in the bottom left image of Figure~\ref{fig:flickr_qualitative} it is possible to detect
that the pedestrian wants to hail a cab and is not intending to cross the road even though she is
standing at the side of the road.

\begin{figure*}[t]
  \centering
  \includegraphics[height=4.27cm,trim=1cm 0.25cm 0.5cm 1.75cm,clip]{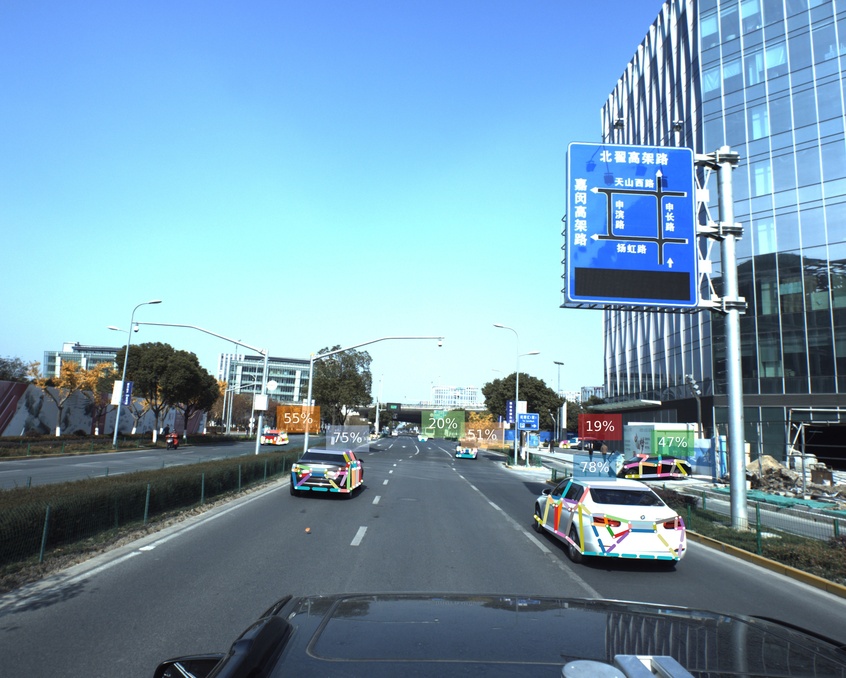}
  \includegraphics[height=4.27cm,trim=0cm 0.25cm 0cm 1.25cm,clip]{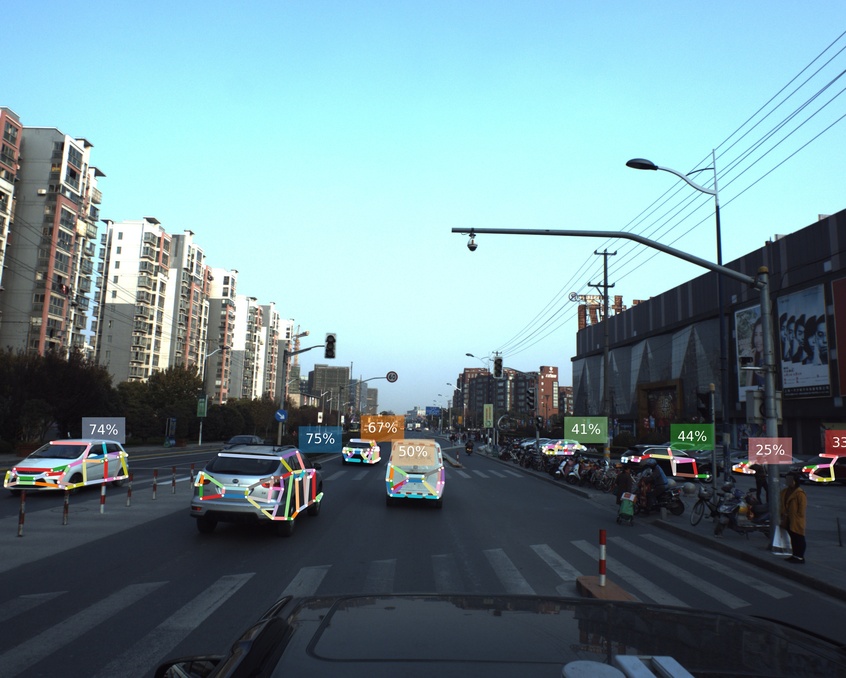}
  \includegraphics[height=4cm,trim=0cm 0.25cm 0cm 1.25cm,clip]{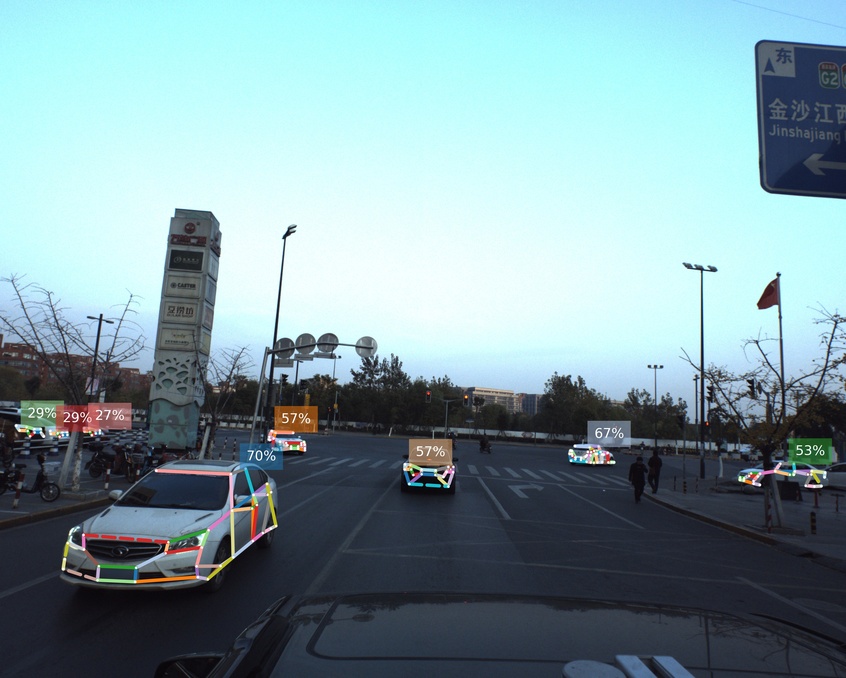}
  \includegraphics[height=4cm,trim=0cm 0.25cm 0cm 1.25cm,clip]{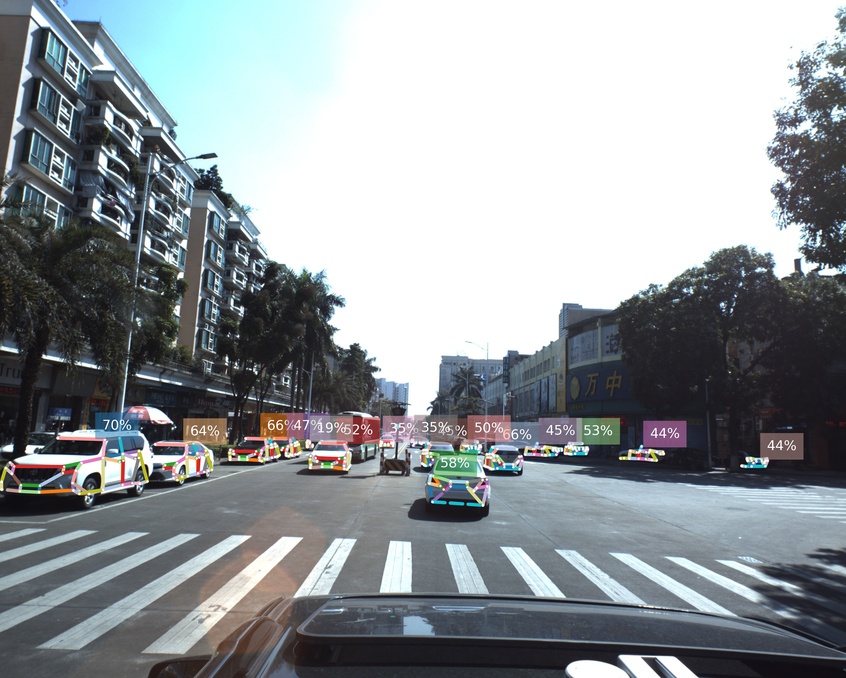}
  \caption{
    Qualitative results from the ApolloCar3D validation set~\cite{song2019apollocar3d}.
    We demonstrate excellent detection rates and spatial localizations of all
    the visible car keypoints even at far distances.
  }
  \label{fig:apollo_qualitative}
\end{figure*}

\paragraph{Results on the ApolloCar3D dataset.}
Our method is not specific to human poses and can be applied to any pose.
To demonstrate that our method generalizes, we apply it to the ApolloCar3D
dataset~\cite{song2019apollocar3d},
where every car instance is annotated with up to 66~keypoints.
Our method achieves an average
precision (AP) of 72.0\% with all sigmas for the computation of the object
keypoint similarity set to 0.05.
The previous work~\cite{song2019apollocar3d} evaluates detection rate instead of AP and a comparison of our method with their Convolutional Pose Machines~\cite{wei2016convolutional} evaluation and
human annotators is shown in Table~\ref{tab:results_apollo}. We achieve a detection rate
of 91.9\% and thus outperform the previous state-of-the-art
from~\cite{song2019apollocar3d} which
achieved a detection rate of 75.4\%.  They also report the detection rate for
human annotators at 92.4\%. Our proposed method reduces the gap
to human level performance from 17.0\% to just 0.5\%.

\begin{table}
  \centering
  \begin{tabular}{|l|c |}
    \hline
    Method & Detection rate[\%] \\
    \hline\hline
    Human annotators & 92.4 \\
    \hline
    CPM~\cite{wei2016convolutional}  & 75.4 \\
    \textbf{Ours}    & \textbf{91.9} \\
    \hline
  \end{tabular}
  \caption{
    Detection rate on the ApolloCar3D dataset~\cite{song2019apollocar3d}. The metrics for the Convolutional Pose
    Machines and the human annotators are from~\cite{song2019apollocar3d}.
  }
  \label{tab:results_apollo}
\end{table}

We share qualitative results on the validation set of
ApolloCar3D~\cite{song2019apollocar3d} in Figure~\ref{fig:apollo_qualitative}.
We predict fine-grained keypoints for close-by and far-away cars.
Cameras for self-driving technology cover a wide angle and therefore have to
perceive small car instances even for cars at moderate distances.
It is safety relevant to determine the locations of break and indicator
lights for downstream tasks which we achieve with high precision.

\begin{table}
  \centering
  \begin{tabular}{|l|c c c c c|}
    \hline
    Method                  & \textbf{WB} & body & foot & face & hand \\
    \hline\hline
    \textbf{Baseline}       & \textbf{55.3} & \textbf{60.8} & \textbf{54.2} & 86.2 & 50.6  \\
    Equal                   & -3.1\% & -4.8\% & -8.1\% & \textbf{+1.5\%} & +0.8\% \\
    Global                  & -1.2\% & -2.5\% & -1.1\% & +0.8\% & \textbf{+2.6\%} \\
    Crafted                 & -1.2\% & -1.6\% & -1.7\% & +1.2\% & -1.0\% \\
    \hline
  \end{tabular}
  \caption{
    Ablation studies.
    Average precision (AP) results in percent on the COCO WholeBody
    val dataset~\cite{jin2020whole}. We report the percentage gain of other weighting
    methods in comparison to our baseline method. In addition to our main method, we
    compare with applying no weighting (Equal), applying our method to the global graph instead of ego graphs (Global) and using hand-crafted weights (Crafted).
    All results are produced with a ShuffleNetV2k16 backbone.
    WB indicates an evaluation on all 133 keypoints.
  }
  \label{tab:ablation}
\end{table}

\paragraph{Ablation Studies.}
We study the effects of different loss weightings on the average precision (AP). In Table~\ref{tab:ablation}
the results from training with different weighting schemes
are shown. For this ablation study we train for 50 epochs starting from a model that was pretrained on the 133 keypoints without weighting. ``Crafted'' means that the body and foot keypoints are weighted
three times higher than the rest of the keypoints. Hand-crafting the weights can be seen
as choosing 133 additional hyperparameters which is generally infeasible and the
motivation for our method.

The weighting based on the local harmonic centrality, which is our baseline method, achieves an AP of 55.3\%,
which is a significant improvement over the unweighted training, which results in an AP of 53.6\%.
The weighting based on the local harmonic centrality also achieves a higher AP than
the weighting based on the vanilla harmonic centrality and the hand-crafted weights. This shows
a) that the local influence between the keypoints is more important than the global one and b) that
using our proposed centrality measure is more optimal than hand-crafting the weights.

\begin{table}
  \centering
   
     \begin{tabular}{|l l| c c c|}
     \hline
     Seed thr. & CAF thr. & AP & t [ms] & t$_{dec}$ [ms]\\
     \hline\hline
     \multicolumn{2}{|c|}{ZoomNet \cite{jin2020whole}} & 54.1 & 175  & - \\
     \hline
     0.2 & 0.001 & \textbf{60.4} & 153 & 60 \\   
     0.5 & 0.001 & 58.4 & 120 & 27 \\   
     0.5 & 0.01 & 54.6 & \textbf{112} & \textbf{20} \\   
     \hline
   \end{tabular}   
  \caption{
    Ablation studies.
    Average precision (AP) results in percent on the COCO WholeBody
    val dataset~\cite{jin2020whole} and their associated prediction time for
    different decoding methods. We show ZoomNet's average precision and runtime for better comparison
    with our method. Our neural network runs in 93ms on a NVIDIA GTX 1080 Ti.
    The decoding starts with the joints
    where the confidence exceeds the seed threshold.
    The vector and scale from a cell of a CAF field
    is only used if the confidence 
    is above the CAF threshold.
  }
  \label{tab:runtime}
\end{table}

We study the effect of different parameter choices
on the precision and prediction time trade-off of our method. The runtime of our method
is influenced by the decoding algorithm. Since the WholeBody pose has 133 joints
and 152 associations the duration of the decoding is more prevalent than for
poses with a lower number of joints and associations.
The first step in the decoding process is to determine seed joints from which the
decoding starts and from which connections to the other joints will be created with the help of the CAF fields.
All joints that have a confidence that is higher than a certain seed threshold will be
used as seed joints. Increasing the seed threshold will reduce the number of seeds
and thus cause a faster decoding process. However, with a higher seed threshold,
some humans may not be detected which can result in a lower average precision.
Using a higher CAF threshold
generally increases the decoding speed in exchange for a lower
accuracy as fewer associations are considered for decoding.
The quantitative results that the variation of these parameters yield can be seen in
Table~\ref{tab:runtime}. With our standard decoder setting, we already achieve a higher AP than ZoomNet~\cite{jin2020whole} whilst being 22~ms faster. With a seed threshold of 0.5 and a CAF threshold of 0.01
our model achieves an AP of 54.5 with a prediction time of 112~ms, which is an excellent trade-off between inference speed and precision.
Our ShuffleNetV2K16 backbone achieves an AP of 50.9 at a frame rate of 15.2 frames per second on a NVIDIA GTX 1080 Ti, which makes at suitable for most real-time applications that require fine-grained pose estimation.




\section{Conclusion}

We have proposed a generic and principled method to train complex poses with fine and
coarse-grained details. Our experiments demonstrate our ability to perceive
detailed facial expressions and hand gestures and
produce state-of-the-art results
on standard pose benchmarks for human and car poses. We have shown that our method
operates at state-of-the-art prediction speeds and we have studied the trade-offs
between accuracy and prediction speeds.

\section{Acknowledgments}
This project has received funding from the Initiative for Media Innovation based at Media Center, EPFL, Lausanne, Switzerland.

{\small
\bibliographystyle{ieee_fullname}
\bibliography{references}

\begin{thebibliography}{10}\itemsep=-1pt

\bibitem{bavelas1950communication}
Alex Bavelas.
\newblock Communication patterns in task-oriented groups.
\newblock {\em The journal of the acoustical society of America},
  22(6):725--730, 1950.

\bibitem{Bertoni2020sociald}
Lorenzo Bertoni, Sven Kreiss, and Alexandre Alahi.
\newblock Perceiving humans: from monocular 3d localization to social
  distancing.
\newblock {\em IEEE Transactions on Intelligent Transportation Systems}, 2021.

\bibitem{bonacich1987power}
Phillip Bonacich.
\newblock Power and centrality: A family of measures.
\newblock {\em American journal of sociology}, 92(5):1170--1182, 1987.

\bibitem{bottou2010large}
L{\'e}on Bottou.
\newblock Large-scale machine learning with stochastic gradient descent.
\newblock In {\em Proceedings of COMPSTAT'2010}, pages 177--186. Springer,
  2010.

\bibitem{cao2019openpose}
Zhe Cao, Gines~Hidalgo Martinez, Tomas Simon, Shih-En Wei, and Yaser~A Sheikh.
\newblock Openpose: realtime multi-person 2d pose estimation using part
  affinity fields.
\newblock {\em IEEE transactions on pattern analysis and machine intelligence},
  2019.

\bibitem{cao2017realtime}
Zhe Cao, Tomas Simon, Shih-En Wei, and Yaser Sheikh.
\newblock Realtime multi-person 2d pose estimation using part affinity fields.
\newblock In {\em Conference on Computer Vision and Pattern Recognition
  (CVPR)}, pages 7291--7299, 2017.

\bibitem{cheng2020higherhrnet}
Bowen Cheng, Bin Xiao, Jingdong Wang, Honghui Shi, Thomas~S Huang, and Lei
  Zhang.
\newblock Higherhrnet: Scale-aware representation learning for bottom-up human
  pose estimation.
\newblock In {\em Proceedings of the IEEE/CVF Conference on Computer Vision and
  Pattern Recognition}, pages 5386--5395, 2020.

\bibitem{dinesh2018carfusion}
N Dinesh~Reddy, Minh Vo, and Srinivasa~G Narasimhan.
\newblock Carfusion: Combining point tracking and part detection for dynamic 3d
  reconstruction of vehicles.
\newblock In {\em Proceedings of the IEEE Conference on Computer Vision and
  Pattern Recognition (CVPR)}, pages 1906--1915, 2018.

\bibitem{freeman1977set}
Linton~C Freeman.
\newblock A set of measures of centrality based on betweenness.
\newblock {\em Sociometry}, pages 35--41, 1977.

\bibitem{he2017mask}
Kaiming He, Georgia Gkioxari, Piotr Doll{\'a}r, and Ross Girshick.
\newblock Mask r-cnn.
\newblock In {\em Computer Vision (ICCV), 2017 IEEE International Conference
  on}, pages 2980--2988. IEEE, 2017.

\bibitem{he2016deep}
Kaiming He, Xiangyu Zhang, Shaoqing Ren, and Jian Sun.
\newblock Deep residual learning for image recognition.
\newblock In {\em Conference on Computer Vision and Pattern Recognition
  (CVPR)}, pages 770--778, 2016.

\bibitem{jin2020whole}
Sheng Jin, Lumin Xu, Jin Xu, Can Wang, Wentao Liu, Chen Qian, Wanli Ouyang, and
  Ping Luo.
\newblock Whole-body human pose estimation in the wild.
\newblock In {\em European Conference on Computer Vision (ECCV)}, 2020.

\bibitem{katz1953new}
Leo Katz.
\newblock A new status index derived from sociometric analysis.
\newblock {\em Psychometrika}, 18(1):39--43, 1953.

\bibitem{ke2020gsnet}
Lei Ke, Shichao Li, Yanan Sun, Yu-Wing Tai, and Chi-Keung Tang.
\newblock Gsnet: Joint vehicle pose and shape reconstruction with geometrical
  and scene-aware supervision.
\newblock In {\em European Conference on Computer Vision (ECCV)}, pages
  515--532. Springer, 2020.

\bibitem{kendall2017uncertainties}
Alex Kendall and Yarin Gal.
\newblock What uncertainties do we need in bayesian deep learning for computer
  vision?
\newblock In {\em Advances in neural information processing systems}, pages
  5574--5584, 2017.

\bibitem{kocabas2018multiposenet}
Muhammed Kocabas, Salih Karagoz, and Emre Akbas.
\newblock Multiposenet: Fast multi-person pose estimation using pose residual
  network.
\newblock In {\em European Conference on Computer Vision (ECCV)}, pages
  417--433, 2018.

\bibitem{kreiss2019pifpaf}
Sven Kreiss, Lorenzo Bertoni, and Alexandre Alahi.
\newblock Pifpaf: Composite fields for human pose estimation.
\newblock In {\em Conference on Computer Vision and Pattern Recognition
  (CVPR)}, June 2019.

\bibitem{kreiss2021openpifpaf}
Sven Kreiss, Lorenzo Bertoni, and Alexandre Alahi.
\newblock {OpenPifPaf: Composite Fields for Semantic Keypoint Detection and
  Spatio-Temporal Association}.
\newblock {\em IEEE Transactions on Intelligent Transportation Systems}, 2021.

\bibitem{lin2017focal}
Tsung-Yi Lin, Priya Goyal, Ross Girshick, Kaiming He, and Piotr Doll{\'a}r.
\newblock Focal loss for dense object detection.
\newblock In {\em Proceedings of the IEEE international conference on computer
  vision (ICCV)}, pages 2980--2988, 2017.

\bibitem{lin2014microsoft}
Tsung-Yi Lin, Michael Maire, Serge Belongie, James Hays, Pietro Perona, Deva
  Ramanan, Piotr Doll{\'a}r, and C~Lawrence Zitnick.
\newblock Microsoft coco: Common objects in context.
\newblock In {\em European Conference on Computer Vision (ECCV)}, pages
  740--755. Springer, 2014.

\bibitem{ma2018shufflenet}
Ningning Ma, Xiangyu Zhang, Hai-Tao Zheng, and Jian Sun.
\newblock Shufflenet v2: Practical guidelines for efficient cnn architecture
  design.
\newblock In {\em European Conference on Computer Vision (ECCV)}, pages
  116--131, 2018.

\bibitem{marchiori2000harmony}
Massimo Marchiori and Vito Latora.
\newblock Harmony in the small-world.
\newblock {\em Physica A: Statistical Mechanics and its Applications},
  285(3-4):539--546, 2000.

\bibitem{mathis2018deeplabcut}
Alexander Mathis, Pranav Mamidanna, Kevin~M Cury, Taiga Abe, Venkatesh~N
  Murthy, Mackenzie~Weygandt Mathis, and Matthias Bethge.
\newblock Deeplabcut: markerless pose estimation of user-defined body parts
  with deep learning.
\newblock Technical report, Nature Publishing Group, 2018.

\bibitem{Mokhtarzadeh2019cvpr}
Sina Mokhtarzadeh, Mina Ghadimi, Ahmad Nickabadi, and Alexandre Alahi.
\newblock Convolutional relational machine for group activity recognition.
\newblock In {\em IEEE Conference on Computer Vision and Pattern Recognition
  (CVPR)}, 2019.

\bibitem{Mordan20its}
Taylor Mordan, Matthieu Cord, Patrick Perez, and Alexandre Alahi.
\newblock Detecting 32 pedestrian attributes for autonomous vehicles.
\newblock {\em IEEE Transactions on Intelligent Transportation Systems - under
  review}, 2020.

\bibitem{nesterov27method}
Yurrii Nesterov.
\newblock A method of solving a convex programming problem with convergence
  rate o(1/k2).
\newblock In {\em Soviet Mathematics Doklady}, volume~27, pages 372--376, 1983.

\bibitem{newell2017associative}
Alejandro Newell, Zhiao Huang, and Jia Deng.
\newblock Associative embedding: End-to-end learning for joint detection and
  grouping.
\newblock In {\em Advances in Neural Information Processing Systems}, pages
  2277--2287, 2017.

\bibitem{newell2016stacked}
Alejandro Newell, Kaiyu Yang, and Jia Deng.
\newblock Stacked hourglass networks for human pose estimation.
\newblock In {\em European Conference on Computer Vision (ECCV)}, pages
  483--499. Springer, 2016.

\bibitem{papandreou2018personlab}
George Papandreou, Tyler Zhu, Liang-Chieh Chen, Spyros Gidaris, Jonathan
  Tompson, and Kevin Murphy.
\newblock Personlab: Person pose estimation and instance segmentation with a
  bottom-up, part-based, geometric embedding model.
\newblock In {\em European Conference on Computer Vision (ECCV)}, pages
  269--286, 2018.

\bibitem{pishchulin2016deepcut}
Leonid Pishchulin, Eldar Insafutdinov, Siyu Tang, Bjoern Andres, Mykhaylo
  Andriluka, Peter~V Gehler, and Bernt Schiele.
\newblock Deepcut: Joint subset partition and labeling for multi person pose
  estimation.
\newblock In {\em Conference on Computer Vision and Pattern Recognition
  (CVPR)}, pages 4929--4937, 2016.

\bibitem{haziq2020}
Haziq Razali and Alexandre Alahi.
\newblock Pedestrian intention prediction: A convolutional bottom-up approach.
\newblock {\em Transportation Research Part C}, 2021.

\bibitem{reddy2019occlusion}
N~Dinesh Reddy, Minh Vo, and Srinivasa~G Narasimhan.
\newblock Occlusion-net: 2d/3d occluded keypoint localization using graph
  networks.
\newblock In {\em Proceedings of the IEEE Conference on Computer Vision and
  Pattern Recognition (CVPR)}, pages 7326--7335, 2019.

\bibitem{sanchez2020simple}
H{\'e}ctor~Corrales S{\'a}nchez, Antonio~Hern{\'a}ndez Mart{\'\i}nez,
  Rub{\'e}n~Izquierdo Gonzalo, Noelia~Hern{\'a}ndez Parra, Ignacio~Parra
  Alonso, and David Fernandez-Llorca.
\newblock Simple baseline for vehicle pose estimation: Experimental validation.
\newblock {\em IEEE Access}, 8:132539--132550, 2020.

\bibitem{shi2016real}
Wenzhe Shi, Jose Caballero, Ferenc Husz{\'a}r, Johannes Totz, Andrew~P Aitken,
  Rob Bishop, Daniel Rueckert, and Zehan Wang.
\newblock Real-time single image and video super-resolution using an efficient
  sub-pixel convolutional neural network.
\newblock In {\em Conference on Computer Vision and Pattern Recognition
  (CVPR)}, pages 1874--1883, 2016.

\bibitem{song2019apollocar3d}
Xibin Song, Peng Wang, Dingfu Zhou, Rui Zhu, Chenye Guan, Yuchao Dai, Hao Su,
  Hongdong Li, and Ruigang Yang.
\newblock Apollocar3d: A large 3d car instance understanding benchmark for
  autonomous driving.
\newblock In {\em Proceedings of the IEEE Conference on Computer Vision and
  Pattern Recognition}, pages 5452--5462, 2019.

\bibitem{sun2019deep}
Ke Sun, Bin Xiao, Dong Liu, and Jingdong Wang.
\newblock Deep high-resolution representation learning for human pose
  estimation.
\newblock In {\em Conference on Computer Vision and Pattern Recognition
  (CVPR)}, pages 5693--5703, 2019.

\bibitem{toshev2014deeppose}
Alexander Toshev and Christian Szegedy.
\newblock Deeppose: Human pose estimation via deep neural networks.
\newblock In {\em Conference on Computer Vision and Pattern Recognition
  (CVPR)}, pages 1653--1660, 2014.

\bibitem{wei2016convolutional}
Shih-En Wei, Varun Ramakrishna, Takeo Kanade, and Yaser Sheikh.
\newblock Convolutional pose machines.
\newblock In {\em Conference on Computer Vision and Pattern Recognition
  (CVPR)}, pages 4724--4732, 2016.

\bibitem{xiang2014beyond}
Yu Xiang, Roozbeh Mottaghi, and Silvio Savarese.
\newblock Beyond pascal: A benchmark for 3d object detection in the wild.
\newblock In {\em Proceeding of the IEEE Winter Conference on Applications of
  Computer Vision (WACV)}, pages 75--82. IEEE, 2014.

\bibitem{xiao2018simple}
Bin Xiao, Haiping Wu, and Yichen Wei.
\newblock Simple baselines for human pose estimation and tracking.
\newblock In {\em Proceedings of the European conference on computer vision
  (ECCV)}, pages 466--481, 2018.

\end{thebibliography}
}

\end{document}